\documentclass[onecolumn,11pt]{article}
\usepackage[top=1in, bottom=1in, left=1in, right=1in]{geometry}
\usepackage{amsmath,amsfonts,amscd,amssymb,amsthm}
\usepackage{graphicx}
\usepackage{epstopdf}
\usepackage{overpic}
\usepackage{cancel}
\usepackage{rotating}
\usepackage{url}
\usepackage{caption}
\usepackage{color}
\usepackage{rotating}
\usepackage{multirow}
\usepackage{wrapfig}
\usepackage{mathtools}
\usepackage{subeqnarray}
\usepackage{setspace}
\usepackage{palatino} 
\usepackage[numbers,sort&compress]{natbib}
\usepackage{bm}
\usepackage{tikz}
\usepackage{enumitem}
\usepackage{tabu}

\theoremstyle{remark}

\usepackage[bottom,flushmargin,hang,multiple]{footmisc}
\usepackage{lipsum}
\newcommand\blfootnote[1]{%
  \begingroup
  \renewcommand\thefootnote{}\footnote{#1}%
  \addtocounter{footnote}{-1}%
  \endgroup
}

\definecolor{header1}{cmyk}{0,0,0,1}

\title{\vspace{-.55in}{\fontsize{16}{16}\selectfont \textbf{Discovering Governing Equations from Partial Measurements with Deep Delay Autoencoders}}\vspace{-.175in}}

\author{\normalsize{Joseph Bakarji$^{1*}$, Kathleen Champion$^{2\dagger}$, J. Nathan Kutz$^2$, Steven L. Brunton$^1$}\\
\footnotesize{$^1$ Department of Mechanical Engineering, University of Washington, Seattle, WA 98195, United States}\\
\footnotesize{$^2$ Department of Applied Mathematics, University of Washington, Seattle, WA 98195, United States\vspace{-.275in}}
}
\date{}
\graphicspath{/figures}
\begin{document}
\maketitle

\blfootnote{$^*$ Corresponding author (jbakarji@uw.edu).}
\blfootnote{$^\dagger$ Work performed independently of employment at Amazon.}
\vspace{-.2in}
\begin{abstract}
A central challenge in data-driven model discovery is the presence of hidden, or latent, variables that are not directly measured but are dynamically important.
Takens' theorem provides conditions for when it is possible to augment these partial measurements with time delayed information, resulting in an attractor that is diffeomorphic to that of the original full-state system.  
However, the coordinate transformation back to the original attractor is typically unknown, and learning the dynamics in the embedding space has remained an open challenge for decades.  
Here, we design a custom deep autoencoder network to learn a coordinate transformation from the delay embedded space into a new space where it is possible to represent the dynamics in a sparse, closed form. 
We demonstrate this approach on the Lorenz, R\"ossler, and Lotka-Volterra systems, learning dynamics from a single measurement variable. 
As a challenging example, we learn a Lorenz analogue from a single scalar variable extracted from a video of a chaotic waterwheel experiment.  
The resulting modeling framework combines deep learning to uncover effective coordinates and the sparse identification of nonlinear dynamics (SINDy) for interpretable modeling.  
Thus, we show that it is possible to simultaneously learn a closed-form model and the associated coordinate system for partially observed dynamics.  \\
\vspace{-.075in}

\vspace{-.05in}
\end{abstract}

\section{Introduction}
Scientific progress has been driven by the discovery of simple and predictive mathematical models from observations.  
Interpretable, parsimonious governing equations have been especially valuable as they typically have allowed for greater engineering insight, simple parametrizations, and improved extrapolation capabilities.  
Increasingly, these models are learned from data using machine learning algorithms, such as genetic programming~\cite{schmidt2009distilling,Bongard2007pnas,Schmidt2011pb,Daniels2015naturecomm} and the sparse identification of nonlinear dynamics (SINDy)~\cite{brunton2016discovering, rudy2017data}. 
When only partial observations of a dynamical system are available, so that some states remain hidden, it is generally not possible to formulate a closed-form model in these variables, using either analytic or data-driven techniques.  
Time-delay embedding provides an approach to augment these limited measurements, and under certain conditions, given by Takens' embedding theorem~\cite{takens1981detecting}, the delay-augmented state yields an attractor that is diffeomorphic (i.e. continuously differentiably deformable) to the underlying, though unmeasured, full-state attractor.  
However, the coordinate transformation back to the original attractor is unknown and may be arbitrarily complex to represent.  
Learning interpretable and generalizable dynamical systems models in the embedding space has remained an open challenge since Takens introduced the embedding theorem in 1981~\cite{takens1981detecting}, with several promising recent developments~\cite{Daniels2015naturecomm,somacal2020uncovering,atkinson2020bayesian,ribera2021model}.  
The central goal of this work is to learn parsimonious, closed-form models from partially observed systems, leveraging the SINDy algorithm for model discovery and a custom deep autoencoder to learn an appropriate coordinate transformation from a time-delay embedding.   

Time-delay embedding has a rich history in data-driven modeling~\cite{roux1983observation,Broomhead1986physd,Crutchfield1987cs,Broomhead1989prsla,kennel1992method,deyle2011generalized,Sugihara2012science,Giannakis2012pnas,Billings2013book,ye2015distinguishing,Ye2015pnas,brunton2017chaos,Loiseau2018jfm,pan2018data,das2019delay,Giannakis2019acha,runge2019inferring,zou2019complex,pan2020structure,giannakis2020delay}, especially for making predictions when analytical models are not available. 
Notable applications of the theory include distinguishing noise from chaotic dynamics~\cite{roux1983observation,kennel1992method}, and recent work has shown that delay embedding can improve machine learning models of time-series data~\cite{lee2019linking, Wehmeyer2018jcp}.
Numerous attempts have been made to learn dynamical systems models directly in the delay embedded space.  
For example, when SINDy was applied to a time-delay embedding for chaotic systems, it was shown that the simplest model that describes the data is linear, establishing a strong connection to Koopman operator theory~\cite{brunton2017chaos}.  
This connection between delay embedding and Koopman has a longer history~\cite{mezic2004comparison,Susuki2015cdc,brunton2017chaos,arbabi2017ergodic,champion2019discovery,das2019delay,Giannakis2019acha,pan2020structure,Kamb2020siads,hirsh2021structured}, including the use of dynamic mode decomposition (DMD)~\cite{Schmid2010jfm,kutz2016dynamic} on time-delay coordinates~\cite{tu2014dynamic, brunton2016extracting}. 
However, for chaotic systems, these linear models in delay coordinates suffer from closure issues and fall short of the goal of parsimonious nonlinear models.  
In related work, attempts have been made to extend SINDy to partially measured systems, using the variational annealing data assimilation technique~\cite{ribera2021model} or applying SINDy to higher derivatives of the measured variables~\cite{somacal2020uncovering}, which is closely related to time-delay embedding.

\begin{figure}[t]
\begin{center}
\includegraphics[width=\linewidth]{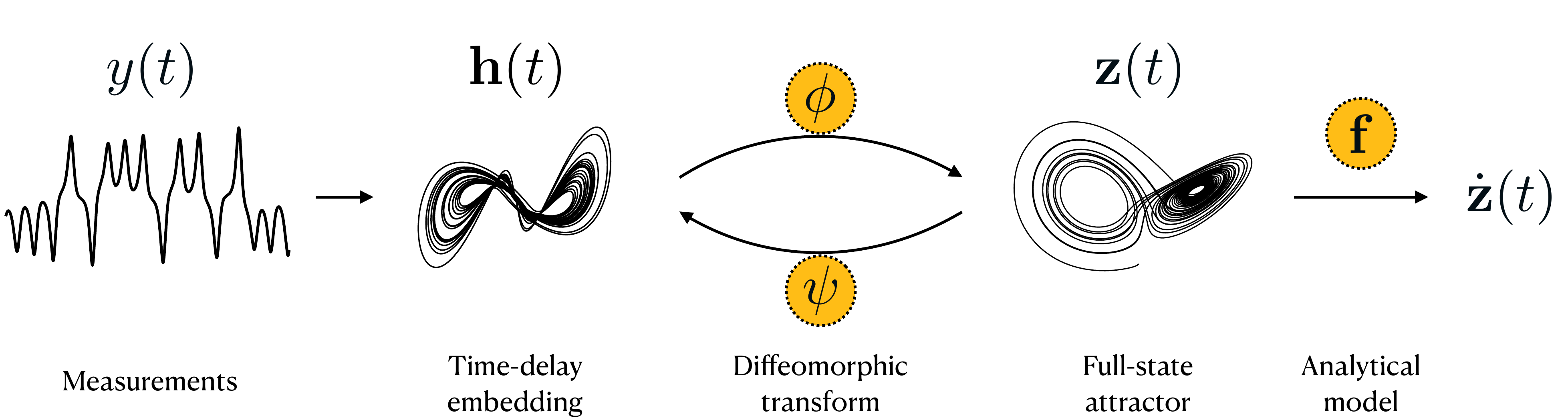}
\caption{Summary of workflow. ${\bm \phi}$, ${\bm \psi}$ and $\mathbf f$ are parametric functions to be discovered.}
\label{fig:workflow}
\end{center}
\end{figure}

Because the data in the delay-embedded space is diffeomorphic to the full-state attractor, where a parsimonious nonlinear model often exists, it is reasonable to try and learn a coordinate transformation starting from a delay-embedded state. 
Figure~\ref{fig:workflow} shows a schematic of the modeling task, involving learning a coordinate transform into a space where an analytical model may be obtained.  
Deep neural network autoencoders have shown great promise in representing arbitrarily complex coordinate transformations, especially for dynamical systems~\cite{Takeishi2017nips,lusch2018deep,Yeung2017arxiv,Mardt2018natcomm,Wehmeyer2018jcp,champion2019data,Otto2019siads,linot2020deep,gilpin2020deep,lee2020model,kalia2021learning}. 
Importantly, Gilpin~\cite{gilpin2020deep} introduced a false nearest neighbors loss function in a deep neural network to promote embedded attractors that are similar to the full-state attractor.  
Similarly, Wehmeyer and No{\'e}~\cite{Wehmeyer2018jcp} showed that autoencoders may be combined with time-delayed measurements to learn effective slow reaction coordinates for molecular kinetics. 
However, neither of these studies learned parsimonious models in the autoencoder coordinates.  
Champion et al.~\cite{champion2019data} showed that it is possible to combine autoencoder networks with SINDy to simultaneously learn a nonlinear coordinate transformation and discover a parsimonious dynamics model, making this a promising technique for learning models in delay-embedded coordinates. 

\begin{figure}[t]
\begin{center}
\includegraphics[width=\textwidth]{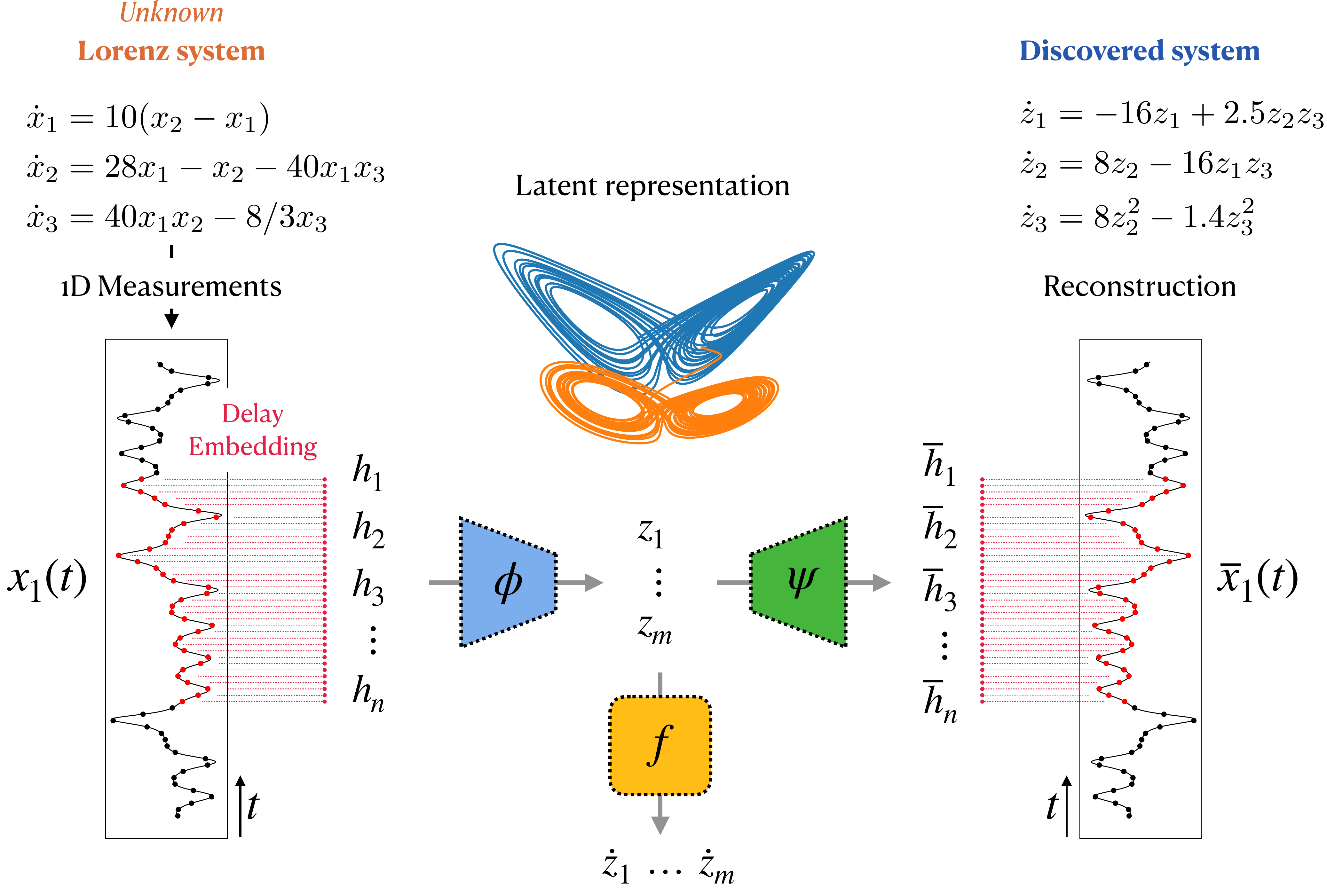}
\caption{General approach to data-driven discovery. ${\bm \phi}$, ${\bm \psi}$ and $\mathbf f$ are the encoder, decoder and analytical model respectively.}
\label{fig:general-approach}
\end{center}
\vspace{-.15in}
\end{figure}

In this work, we leverage the SINDy-autoencoder framework~\cite{champion2019data} to learn parsimonious governing dynamical systems from delay-embedding of partial observations, as summarized in Fig.~\ref{fig:general-approach}. 
In general, model discovery from full-state measurements is an ill-posed inverse problem, and model discovery from limited measurements compounds this challenge.  
As with other custom autoencoders that are used to learn dynamics~\cite{lusch2018deep,champion2019data}, there are several critical steps, including choosing the architecture, designing the loss function, initializing the parameters, and tuning the optimization algorithm used to train the model.  
In addition to demonstrating this algorithm to learn closed-form models from partial observations of several dynamical systems, we describe how this inverse modeling problem may be broken down into sub-problems.  
In general, starting with a problem where the answer is known is important to guide the design of the architecture and loss functions.  
Thus, we show how the architecture and individual terms in the loss function affect the learning process with different levels of assumed knowledge.  
We also investigate what preprocessing should be applied to the data, such as whether or not to apply the singular value decomposition to the delay coordinates before the autoencoder network.  
Incredibly, we show that for the Lorenz system, from a partial observation of only the first coordinate, it is possible to learn a coordinate system in which the dynamics may be represented by a differential equation with even fewer terms than the original Lorenz system; the original system has 7 linear and quadratic terms, while our model only has 6 terms.  
We also show that it is possible to learn a Lorenz analogue from a video sequence of a chaotic waterwheel experiment.  

The rest of the paper is organized as follows.  First, we provide a brief problem definition in Sec.~\ref{Sec:ProblemDefinition}. 
Section~\ref{sec:background} has background information about SINDy and how it has been combined with autoencoders.  
Section~\ref{Sec:DelaySINDy} discusses the proposed delay SINDy autoencoder, including the architecture and custom loss functions. 
The main results from numerical experiments are provided in Sec.~\ref{sec:results}.  We analyze several canonical nonlinear and chaotic systems, in addition to video data from a chaotic waterwheel experiment that may be viewed as a Lorenz analogue.  
Finally, conclusions and discussions are provided in Sec.~\ref{sec:discussion}

\subsection{Mathematical problem definition}\label{Sec:ProblemDefinition}
An overarching inverse problem in data-driven dynamical systems is to learn a model for the evolution of a system, purely from observation data. 
Mathematically, this inverse problem can be stated as follows. 
Let $\mathbf x(t)$, an $m$ dimensional time-dependent state vector defined on the domain $\mathcal D \in \mathbb R^m \times \mathbb R^+$, with $m \in \mathbb N$ and $t > 0$, be the solution of a nonlinear differential equation
\begin{equation}\label{eq:general-eqn}
    \frac{d }{dt} \mathbf x(t)= \mathbf f(\mathbf x(t); \bm \mu)
\end{equation}
where $\mathbf f$ is a smooth and nonlinear function and $\bm \mu$ a vector of system parameters. 
Further, let $\mathbf{y}\in\mathbb{R}^d$ represent noisy measurements of the system, given by
\begin{equation}\label{eq:measurement}
\mathbf{y}(t)=\mathbf{g}(\mathbf{x}(t)) + {\bm \eta},
\end{equation}
where ${\bm \eta}$ is a noise process.  
The goal is then to learn an approximate dynamical system
\begin{equation}\label{eq:reconstructeddyn}
    \frac{d }{dt} \mathbf z(t)= \check{\mathbf f}(\mathbf z(t); \check{\bm \mu})
\end{equation}
in terms of a new state $\mathbf{z}$, which may either be the measured state $\mathbf{y}$ or an invertible function of $\mathbf{y}$:
\begin{equation}\label{eq:embedding}
\mathbf{z}(t)={\bm \phi}(\mathbf{y}(t)).
\end{equation}

If observation data for the full state $\mathbf x(t)$ are available, so that $\mathbf{y}=\mathbf{x}$ and $\mathbf{z}=\mathbf{y}$, an approximation of $\mathbf f$ can be inferred using a variety of machine learning algorithms.  
Although neural networks have been shown to provide useful predictive models~\cite{pathak2018model,vlachas2018data,wu2018deep,raissi2019physics,mardt2020deep}, they don't result in interpretable, closed-form expressions such as \eqref{eq:reconstructeddyn} that are amenable to analysis.  
Parsimonious modeling techniques, such as genetic programming~\cite{schmidt2009distilling,Bongard2007pnas} and SINDy~\cite{brunton2016discovering, rudy2017data}, are able to learn analytic, closed-form models that often capture the exact form of the original dynamics \eqref{eq:general-eqn} when they are known.  

However, in many real world applications, only partial measurements are available, so the dimension of $\mathbf{y}$ is less than the dimension of $\mathbf{x}$, i.e. $d< m$. 
In general, it is not possible to learn $\mathbf f$ directly in its original dimension in terms of the measurement vector $\mathbf{y}$.   
A direct approach of fitting a recurrent relation for $\mathbf{y}$, using recurrent neural networks, Bayesian methods, Gaussian process regression, etc., or discovering the underlying differential equation using SINDy, do not generalize well if the dynamics of $\mathbf{y}$ are nonlinear and depend on hidden variables that control its evolution.  
The challenge of representing the dynamics in terms of a partial observation $\mathbf{y}$ of the full-state $\mathbf{x}$ is known as the \emph{closure} problem, and it is one of the most fundamental issues in scientific modeling, with a particularly rich history in turbulence modeling~\cite{ling2016reynolds,pan2018data,duraisamy2019turbulence,brunton2020machine}. 

There are several embedding techniques~\cite{Coifman2008mmas,Yair2017pnas,brunton2017chaos,gilpin2020deep} to enrich the partial measurement vector, providing a map as in \eqref{eq:embedding}, although it remains a challenge to learn the dynamics \eqref{eq:reconstructeddyn} in these coordinates. 
Therefore, in general, it is necessary to simultaneously learn the coordinate transformation \eqref{eq:embedding} and the dynamics \eqref{eq:reconstructeddyn}. 
For simplicity, we will assume scalar measurements $y(t)$, although the following arguments generalize to multi-dimensional vectors of partial measurements. 
Before learning the map \eqref{eq:embedding}, we first time-delay embed the measurement $y(t)$, resulting in the intermediate vector $\mathbf{h}(t; n, \tau) = [y(t), y(t+\tau), y(t+2\tau), \ldots, y(t +(n-1)\tau)]$, where $n$ and $\tau$ are the number of delays and the time increments between successive measurements respectively.  
Takens' theorem~\cite{takens1981detecting} provides conditions for when time-delay embedding results in an attractor that is diffeomorphic to the original system. 
We then simultaneously learn the map \eqref{eq:embedding} from $\mathbf{z}={\bm \phi}(\mathbf{h})$ and the dynamics \eqref{eq:reconstructeddyn} using a modification of the SINDy-autoencoder framework~\cite{champion2019discovery}.  
The goal is an interpretable differential equation $\dot{\mathbf z}(t) = \check{\mathbf f}(\mathbf z(t); \check{\bm \mu})$ in the transformed embedded variable. 
Takens' embedding theorem also puts a condition on the embedding dimension to be $n>2m$ for the diffeomorphism to be guaranteed.

\section{Background}\label{sec:background}
In this section, we review time-delay embedding, the sparse identification of nonlinear dynamics (SINDy)~\cite{brunton2016discovering} algorithm, and the recent combination of SINDy with deep autoencoders~\cite{champion2019discovery}.

\paragraph{Delay embedding.} When defining the delay embedding variable $\mathbf{h}(t; n, \tau)$ of a measurement time series $\tilde{\mathbf y}$, the proper choice of $\tau$ and $n$ have been studied extensively~\cite{ma2006selection, kennel1992determining} (see appendix \ref{app:embedding-hyper}). 
The embedded variable $\mathbf h(t)$ can be assembled into a Hankel matrix
\begin{equation}\label{eq:hankel}
    \mathbf H = 
\begin{bmatrix}
y(t_1) & y(t_2)& \ldots & y(t_{q})\\
y(t_2) & y(t_3) & \ldots & y(t_{q+1})\\
\vdots & \vdots & \ddots & \vdots \\
y(t_n) & y(t_{n+1}) & \ldots & y(t_{n+q+1})
\end{bmatrix}
=
\begin{bmatrix}
\mathbf h_1,
\mathbf h_2,
\ldots ,
\mathbf h_q
\end{bmatrix}.
\end{equation}
The corresponding eigen-time-delays are given by the columns of $\mathbf{U}$ from the singular value decomposition (SVD) $\mathbf H = \mathbf U \bm \Sigma \mathbf V^\top$, and they are used for linear system identification~\cite{Juang1985jgcd,Juang1991nasatm,Phan:1992,Juang1994book,Brunton2019book} and in the Hankel alternative view of Koopman (HAVOK) algorithm~\cite{brunton2017chaos}. 
The first few eigen-time-delays, whose rank can be chosen based on a Pareto front analysis, are often sufficient to encode the shape of a given signal, as shown in Fig.~\ref{fig:hankel}. 

\paragraph{SINDy.} The SINDy algorithm frames the problem of discovering the model in~\eqref{eq:reconstructeddyn} as a sparse generalized linear regression~\cite{brunton2016discovering}. 
SINDy assumes a library of candidate functions $\bm{\theta}_i(\mathbf z)$ that determine possible terms in $\check{\mathbf f}$.
Given $q$ discrete-time samples of $\mathbf z(t) \in \mathbb R^m$ and $\dot{\mathbf{z}}(t) \in \mathbb R^m$, these may be cast into data matrices $\mathbf Z = [\mathbf z_1, \mathbf z_2, \ldots, \mathbf z_q]^\top$ and $\dot{\mathbf{Z}} = [\dot{\mathbf{z}}_1, \dot{\mathbf{z}}_2, \ldots, \dot{\mathbf{z}}_q]^\top$, where $\mathbf Z$, $\dot{\mathbf{Z}} \in \mathbb R^{q \times m}$.  
Assuming $r$ candidate functions, with $r \ll m$, it is possible to build the library data matrix $\bm \Theta(\mathbf Z) = [\bm \theta_1(\mathbf Z), \bm \theta_2(\mathbf Z), \ldots, \bm \theta_r(\mathbf Z)] \in \mathbb R^{m \times r}$. 
The SINDy algorithm then minimizes the loss
\begin{equation}\label{eq:sindy-loss}
    \mathcal L_\text{SINDy}(\bm \Xi) = \left\|\dot{\mathbf{Z}} - \bm \Theta(\mathbf Z)  \bm \Xi \right\|^2_2 + \lambda \left\| \bm \Xi \right\|^2_0,
\end{equation}
where the sparse matrix $\bm \Xi = [ \bm \xi_1, \bm \xi_2, \ldots, \bm \xi_r] \in \mathbb R^{r \times m}$ determines which candidate terms are active in the dynamics, so that $\check{\mathbf{f}}(\mathbf{z})\approx {\bm \Theta}(\mathbf{z}){\bm \Xi}$.  
The sparsity constraint $\left\| \bm \Xi \right\|^2_0$ minimizes the number of non-zero terms in $\bm \Xi$, but it requires an intractable combinatorial search. 
Thus, it is either replaced by $\left\| \bm \Xi \right\|^2_1$ or a sequentially thresholded least-square algorithm for computational efficiency. 
A sparse regularizer is motivated by Occam's razor, which prefers simpler and more interpretable models.
SINDy has been widely applied to model a range of complex systems~\cite{Schaeffer2017prsa,boninsegna2018sparse,Kaiser2018prsa,Loiseau2017jfm,Brunton2019book,Gelss2019mindy,gurevich2019robust,beetham2020formulating,Reinbold2020pre,schmelzer2020discovery,suri2020capturing,beetham2021sparse,reinbold2021robust}.

\begin{figure}[t]
\begin{center}
\includegraphics[width=15cm]{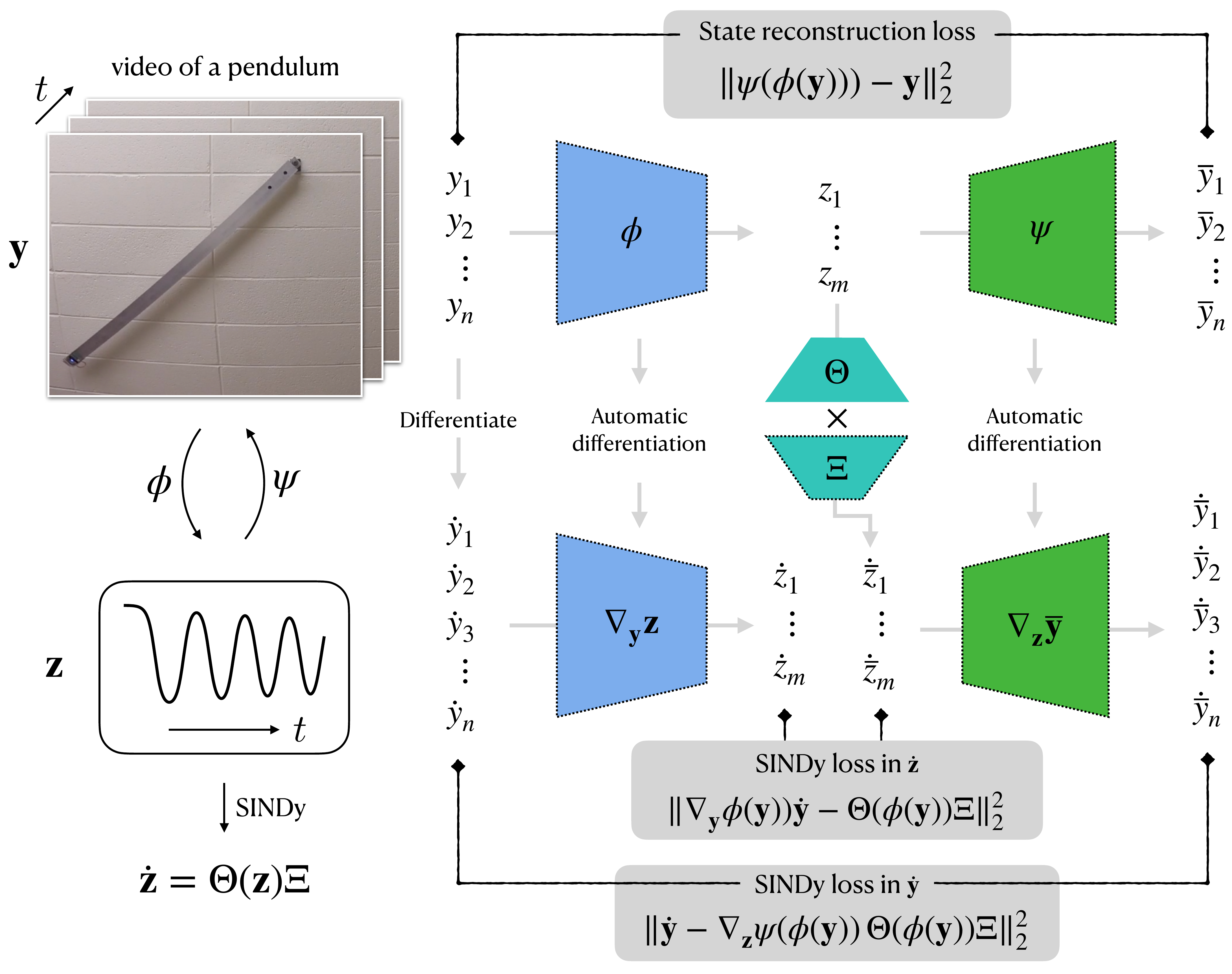}
\vspace{-.2in}
\caption{SINDy Autoencoder summary algorithm. ${\bm \phi}$ and ${\bm \psi}$ are encoder and decoder networks respectively. The network Jacobians $\partial \mathbf z/\partial \mathbf y$ and $\partial \bar{\mathbf{y}}/\partial \bar{\mathbf{z}}$ are computed using automatic differentiation in tensorflow (i.e. \texttt{tf.jacobian}).}
\label{fig:sindy-auto}
\end{center}
\end{figure}

\paragraph{SINDy autoencoders.} SINDy is most effective when the choice of the dictionary is informed by the structure of the problem. 
When the potential terms of the differential equations we seek are not known a priori, or if they include integral operators (e.g. non-local closures~\cite{bakarji2021data}), a naive implementation of SINDy gives sub-optimal results. 
A particularly versatile solution is to use autoencoders to transform the coordinates of the input variable before discovering a differential equation, as in Champion et al.~\cite{champion2019discovery}. The resulting SINDy autoencoder algorithm is summarized in Fig.~\ref{fig:sindy-auto}. 
We will expand on this algorithm to address the problem of data-driven discovery of high-dimensional equations from low-dimensional measurements.

SINDy autoencoders simultaneously transform measurement data $\mathbf y(t) \in \mathbb R^{p}$ into a lower dimensional variable $\mathbf z(t) \in \mathbb R^m$ and discover a nonlinear differential equation $\dot{\mathbf z} = \bm \Theta (\mathbf z) {\bm \Xi}$ in $\mathbf{z}$. The main element of the algorithm is an autoencoder that requires a reconstruction loss
\begin{equation}
    \mathcal L_\text{recon} = \left\| {\bm \psi}( {\bm \phi}(\mathbf y)) - \mathbf y \right\|_2^2.
\end{equation}

In addition, the method incorporates the SINDy loss given in equation~\eqref{eq:sindy-loss} in two different ways. First, the difference between the output of the SINDy layer, $\dot{\bar{\mathbf{z}}} = \bm \Theta(\mathbf z) \bm \Xi = \bm \Theta({\bm \phi}(\mathbf y)) \bm \Xi$, and that computed by the chain rule $\dot{\mathbf{z}} = \nabla_\mathbf{y} {\bm \phi}(\mathbf y) \dot{\mathbf{y}}$, is minimized. This SINDy loss in $\dot{\mathbf{z}}$ is given by
\begin{equation}\label{eq:sindyauto-zdot}
    \mathcal L_{\dot{\mathbf{z}}} = \left\| \nabla_\mathbf{y} {\bm \phi}(\mathbf y) \dot{\mathbf{y}} - \bm \Theta({\bm \phi}(\mathbf y)) \bm \Xi \right\|_2^2.
\end{equation}
Second, a consistency loss ensuring that $\dot{\mathbf{y}}$ can be reconstructed from $\mathbf y$ using the path $\mathbf y \rightarrow \mathbf z \rightarrow \dot{\bar{\mathbf{z}}} \rightarrow \dot{\bar{\mathbf{y}}}$ is included. Using the fact that
\begin{align*}
   \dot{\bar{\mathbf{y}}} 
   &= \frac{\partial \bar{\mathbf{y}} }{ \partial \bar{\mathbf{z}} } \cdot \frac{ \partial \bar{\mathbf{z}} }{ \partial \mathbf t } \\ 
   &= \nabla_\mathbf{z}{\bm \psi}(\mathbf{z}) \cdot \bm \Theta(\mathbf z) \bm \Xi \\
   &= \nabla_\mathbf{z}{\bm \psi}({\bm \phi}(\mathbf y)) \cdot \bm \Theta({\bm \phi}(\mathbf y)) \bm \Xi,
\end{align*}
the resulting loss in $\dot{\mathbf{y}}$ is given by
\begin{equation}\label{eq:sindyauto-ydot}
    \mathcal L_{\dot{\mathbf{y}}} =  \left\| \dot{\mathbf{y}} - \nabla_\mathbf{z}{\bm \psi}({\bm \phi}(\mathbf y)) \bm \Theta({\bm \phi}(\mathbf y)) \bm \Xi \right\|_2^2.
\end{equation}
Finally, a sparsity constraint $\mathcal L_\text{sparse} = \| \bm \Xi \|_1$ is included in the loss function to encourage the discovery of a parsimonious model. 
While the choice of these losses is sufficient for some high-dimensional data, the current problem requires more loss terms to account for delayed inputs. 

\section{Delay SINDy Autoencoders}\label{Sec:DelaySINDy}
We now leverage the SINDy autoencoders algorithm to learn coordinates and dynamics starting from a delay-embedding.  
The algorithm is extended by adding losses that take advantage of time-delay embedding properties.

\begin{figure}[t]
\begin{center}
\includegraphics[width=.775\textwidth]{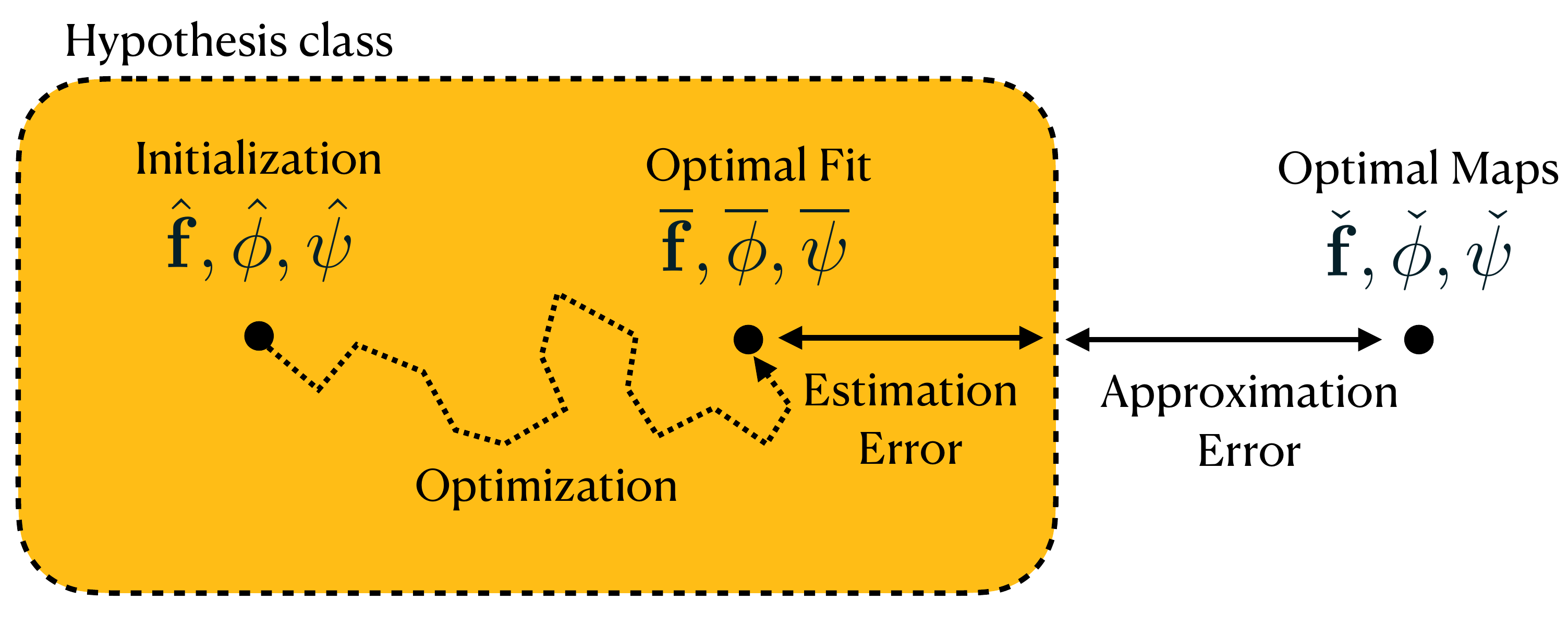}
\caption{Illustration of the estimation and approximation errors in the hypothesis class.}
\label{fig:hyp-class}
\end{center}
\end{figure}

Let $\{\mathbf h_i\}_{i=1}^q$ be the embedding of discrete time-series data $\hat{\mathbf y}=[y(\tau), y(2\tau), y(3\tau), \cdots]$, given by the column vectors of the Hankel matrix in \eqref{eq:hankel}. 
Accordingly, $h_{ij}$ is the $j$-th entry of the embedding vector $\mathbf h_i$. 
Our approach relies on the following premises:
\begin{enumerate}
    \item Takens' embedding theorem proves that a diffeomorphic map between the delay-embedded attractor $\mathbf h$ and the original attractor $\mathbf x(t)$ exists under certain conditions. The theorem doesn't specify how to find this map.
    \item The universal approximation properties of neural networks~\cite{csaji2001approximation} can approximate this diffeomorphism and others where the dynamics are simplified.
\end{enumerate}
The goal is then to design a loss function whose optimal solution is a predictive model with a sparse analytical form. 
Ideally, the optimal sparse differential equation of the latent variable, $\mathbf z(t)$, is similar to that of $\mathbf x(t)$ when it is known.
Finding the best loss function is matter of making the hypothesis class as small as possible and as close as possible to the optimal mappings in such a way to minimize the approximation and estimation errors. This is illustrated in Fig.~\ref{fig:hyp-class}. We explore these premises in Sec.~\ref{sec:results}.

\begin{figure}[t]
\vspace{0.1in}
\begin{center}
\includegraphics[trim=23 0 22 0, clip,width=\textwidth]{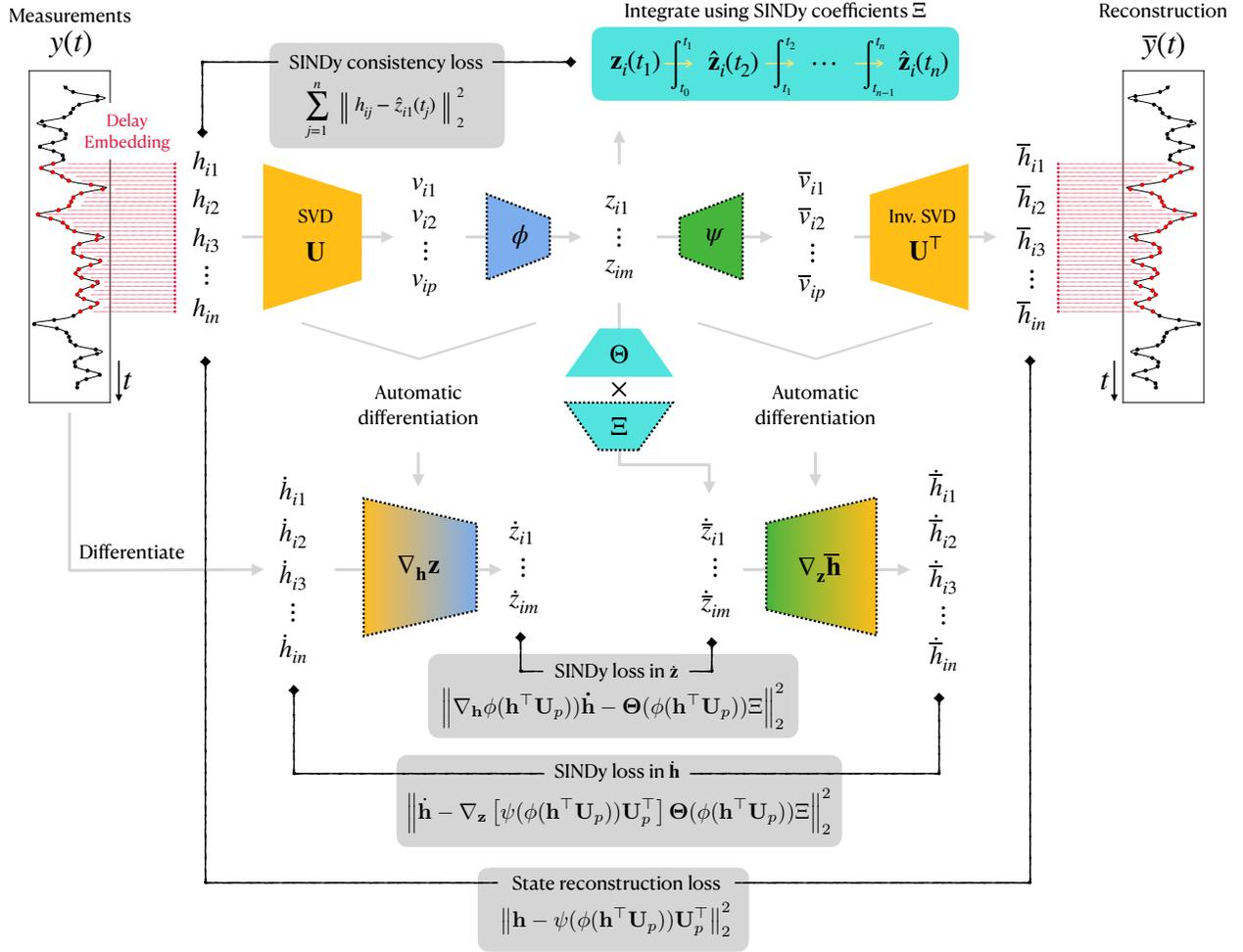}
\caption{Summary of delay SINDy Autoencoders algorithm. The tensor $h_{ij}$ denotes the $i$ row and $j$ column of the Hankel matrix $\mathbf H$.}
\label{fig:sum-alg}
\end{center}
\vspace{0.2in}
\end{figure}

In addition to the SINDy autoencoder losses in Sec.~\ref{sec:background}, we introduce three losses that take advantage of a priori knowledge of time-delay embedding properties, as illustrated in Fig.~\ref{fig:sum-alg}.

\paragraph{Loss 1.}
First, the dimensionality reduction of $\mathbf h$ can be performed using linear techniques, such as computing the SVD of the Hankel matrix in \eqref{eq:hankel}, or using nonlinear techniques, such as a deep autoencoder network. 
The diffeomorphism between $\mathbf h$ and $\mathbf x$ is expected to be highly nonlinear, which motivates the use of neural network. 
However, it is possible to perform an initial linear dimensionality reduction, using the first $p$ dominant modes of the Hankel matrix as an input to the autoencoder.  
This step can filter out noise and improve the computational efficiency of the method. 
If $n\gg2m$, then a linear projection of $\mathbf h$ on an intermediate dimension $p$ preserves the majority of the variance in the signal. 

In this case, the SVD projection $\mathbf h^\top \mathbf U_p$ and its inverse are included as part of the encoding and decoding of the time-series $y(t)$. 
We define $\mathbf U_p$ to be the first $p$ columns of $\mathbf U$, and $\bm \Sigma_p$ the first $p$ rows and columns of $\bm \Sigma$ from the SVD of $\mathbf{H}$. 
The SINDy losses in $\dot{\mathbf{z}}$ and $\dot{\mathbf{y}}$, given in the original SINDy autoencoder algorithm by \eqref{eq:sindyauto-zdot} and \eqref{eq:sindyauto-ydot} respectively, now become
\begin{subequations}
\begin{align}
    \mathcal L_{\dot{\mathbf z}} &= \left\| \nabla_\mathbf{h} \bm \phi(\mathbf h^\top \mathbf U_p  )) \dot{\mathbf h} - \bm \Theta({\bm \phi}(\mathbf h^\top \mathbf U_p )) \bm \Xi \right\|_2^2 \\
    \mathcal L_{\dot{\mathbf h}} &= \left\| \dot{\mathbf h} - \nabla_\mathbf{z} \left[ \bm \psi(\bm \phi(\mathbf h^\top \mathbf U_p )) \mathbf U_p^\top \right] \bm \Theta( \bm \phi(\mathbf h^\top \mathbf U_p  )) \bm \Xi \right\|_2^2,
\end{align}
\end{subequations}
where $\mathbf y$ has been replaced by $\mathbf h$ for consistency. 
Furthermore, the reconstruction loss is obtained from the difference between the feed-forward prediction $\overline{\mathbf h}$ and the input $\mathbf h$
\begin{equation}\label{eq:auto-loss}
    \mathcal L_\text{recon} = \left\| \mathbf h - {\bm \psi}({\bm \phi}(\mathbf h^\top \mathbf U_p  )) \mathbf U_p^\top  \right\|_2^2.
\end{equation}

\paragraph{Loss 2.} Second, we require the first component of the discovered latent variable $\mathbf z(t)$ to be an exact reconstruction of the input signal $y(t)$. This significantly constrains the hypothesis class by an $m$-fold reduction of the number of permutations of the components of $\mathbf z$, to $m-1!$ possibilities instead of $m!$. 
This reduction ultimately makes the optimization problem more feasible. The first component loss is simply given by
\begin{equation}\label{eq:z1loss}
    \mathcal L_{z_1} = \left\| h_{i1} - z_{i1} \right\|^2_2,
\end{equation}
where $\mathbf h_i$ is the $i$-th delay embedding of $\hat{\mathbf y}$ and $\mathbf z_i= [z_{i1}, z_{i2}, \ldots, z_{im}]$ is the corresponding latent variable transformation.

In addition to favorably constraining the optimization, this loss function promotes a more physically interpretable solution.  
The variable $y(t)$ typically has some physical meaning, as it is the quantity that is actually measured experimentally. 
Thus, fixing the first component of $\mathbf{z}$ to be equal to $y(t)$ guarantees that the dynamics in the latent space are directly relevant for the measured quantity $y$. 

\paragraph{Loss 3.} Finally, if the constraint in equation \eqref{eq:z1loss} is satisfied, the first dimension of the time integration of $\mathbf z_i$ using the SINDy model, $\dot{\mathbf z}_i = \bm \Theta(\mathbf z_i) \bm \Xi$, should correspond to the vector $\mathbf h_{i}$. 
In Fig.~\ref{fig:sum-alg}, we set $\hat{\mathbf z}_i(t_1) \equiv \mathbf z_{i}$ as an initial condition where $t_1$ is the initial time, and integrate it using the SINDy model with a time step $\Delta t = t_{i} - t_{i-1}$ that matches the sampling rate of the input data $\hat{\mathbf y}$. The integration can be expressed as
\begin{equation}
    \hat{\mathbf z}_i(t_i) = \int_{t_1}^{t_i} \bm \Theta({\bm \phi}( \mathbf h_i^\top \mathbf U_p)) \bm \Xi \, dt.
\end{equation}
The corresponding SINDy consistency loss is therefore given by
\begin{equation}\label{eq:integral-loss}
    \mathcal L_\text{cons} = \sum_{j=1}^{n} \left\| h_{ij} -  \left(\int_{t_1}^{t_j} \bm \Theta({\bm \phi}( \mathbf h_i^\top \mathbf U_p)) \bm \Xi dt\right)_1 \right\|_2^2,
\end{equation}
where the subscript $1$ of the parentheses designates the first component of $\hat{\mathbf z}(t_i)$, and $t_j = t_1 + j \Delta t $ for $j \in [1..n-1]$. The initial condition $\hat{\mathbf z}(t_1)$ is not included in the loss to avoid redundancy with the $\mathcal L_{z_1}$ loss. The integration procedure is included in the network as a recurrent layer with an RK4 integration scheme.

Finally, all the losses are combined to give
\begin{equation}
    \mathcal L = \mathcal L_\text{recon} + \lambda_1 \mathcal L_{\dot{\mathbf{h}}} + \lambda_2 \mathcal L_{\dot{\mathbf{z}}} + \lambda_3 \mathcal L_{z_1} + \lambda_4 \mathcal L_{\text{cons}} + \lambda_5 \mathcal L_\text{reg},
\end{equation}
with weighting coefficients $\bm \lambda = [\lambda_1, \lambda_2, \ldots, \lambda_5]$ that are hyperparameters to be optimized.

The proposed algorithm constrains the hypothesis class to make the inverse problem computationally feasible and the corresponding optimal solution predictive of the measurement data. However, the choice of the weights $\bm \lambda$ and the losses to include are not unique. Given more information about the system, such as conservation of momentum or energy, more losses can be included to guide the optimization algorithm to find the optimal transform within the hypothesis class.

\newpage
\section{Results}\label{sec:results}
We demonstrate the ability of the proposed deep delay autoencoder network to discover coordinates and parsimonious models for a number of canonical dynamical systems.  
Simultaneously learning the coordinates and dynamics is a challenging inverse problem.  
To solve this inverse modeling problem, and to give insight into potential pitfalls, we break it down into the following sub-problems, of increasing difficulty:
\begin{enumerate}
    \item Optimize ${\bm \phi}$ and ${\bm \psi}$ (with and without an intermediate SVD layer) with known $\mathbf x$; i.e. using supervised learning.  \label{subprob:1} 
    \item Optimize ${\bm \phi}$ and ${\bm \psi}$ (with and without an SVD layer) with known analytical model $\mathbf f$, by fixing the coefficient matrix ${\bm \Xi}$. \label{subprob:2}
    \item Optimize ${\bm \phi}$, ${\bm \psi}$, and ${\bm \Xi}$, initializing $\mathbf f$ in the proximity of its original value in a known system of equations. \label{subprob:3}
    \item Optimize ${\bm \phi}$, ${\bm \psi}$, and ${\bm \Xi}$ with random initialization. \label{subprob:4}
\end{enumerate}
The challenge of solving these optimization problems, in increasing order of difficulty, is due to the size of the hypothesis class which is spanned by the fitting parameters of ${\bm \phi}$, ${\bm \psi}$, and ${\bm \Xi}$, as illustrated in Fig.~\ref{fig:hyp-class}. 
Increasing the complexity of the hypothesis decreases the approximation error by capturing the nonlinear diffeomophisms ${\bm \phi}$ and ${\bm \psi}$. 
In the chaotic systems examples used in this section, we expect $\mathbf f$ to be nonlinear, with at least second order polynomial features. 
However, expanding the hypothesis class makes the optimization task more difficult, thus increasing the estimation error that arises because of sub-optimal fitting (e.g. local minimum, over-fitting, lack of data etc.). 
Accordingly, by assuming some fitting functions to be known prior to optimization, we slowly increase the size of the hypothesis class to explore its robustness and generalization.

We apply the full algorithm to the following systems: the Lorenz system, the Lotka-Volterra equations, the R\"ossler attractor, and a video of the Lorenz waterwheel. 
In the following examples, the first component of $\mathbf x$ is used for the partial measurement $y(t)$.
However, not all scalar measurements $y(t)$ of $\mathbf x$ are equally fit to discover the mappings ${\bm \psi}$ and ${\bm \phi}$. 
For example, measurement of the third component of the Lorenz system does not satisfy Takens' embedding theory, and therefore, the delay-embedded attractor is not diffeomorphic to the full-state attractor $\mathbf x$. 
Furthermore, the original full-state system may not be the only coordinate system that admits a sparse representation of the dynamics, and indeed there may be even sparser representations, as we will show for the Lorenz system. 
Therefore, the maps ${\bm \phi}$ and ${\bm \psi}$ and the dynamics $\check{\mathbf{f}}$ may not be unique, which we will discuss in the following test cases.  

The delay embedding dimension $n$ and the time step $\tau$ are chosen to satisfy the equality $n\tau \approx 0.1$, such that the delay embedding of $y(t)$ is most unfolded in the embedding phase space (see Appendix \ref{app:embedding-hyper}); this choice is also consistent with that from HAVOK~\cite{brunton2017chaos}. 
Along with $n$ and $\tau$, the hyperparameters over which we optimize ${\bm \phi}$, ${\bm \psi}$, and $\mathbf \Xi$ are the SVD rank dimension $p$, the dimension of the latent variable $m$, the architecture of the neural network (number of nodes, number of layers, activation function etc.), and the SINDy library (polynomial order, trigonometric features, threshold value, etc.). 
The high dimensionality of the hyperparameter space adds to the challenge of the optimization problem, which we address in the current study by running a hyperparameter search on parallel GPU clusters.

\subsection{The Lorenz System}
As a benchmark problem, we investigate the three dimensional Lorenz system, which arises in atmospheric modeling and is often used to study chaos and nonlinear dynamics. 
The system of equations for $\mathbf x(t)$ is given by
\begin{subequations}
\begin{align}
    \dot x_1 &= \sigma(x_2 - x_1) \\
    \dot x_2 &= x_1(\rho - x_3) - x_2 \\
    \dot x_3 &= x_1 x_2 - \beta x_3,
\end{align}
\end{subequations}
where one typically assumes $\sigma$, $\rho$, and $\beta$ are positive. The system is known to be chaotic for $\sigma = 10$, $\rho=28$, and $\beta=8/3$. 
In this experiment, we use simulation data of the first component $x_1$ as the measurement $y(t)$, which is input to the neural network.  The goal is to find a coordinate transformation from the delay-embedded coordinates $\mathbf{h}$ to a new coordinate system $\mathbf{z}$ where the dynamics have a sparse, closed-form representation.  

\begin{figure}[t]
\begin{center}
\includegraphics[width=16cm]{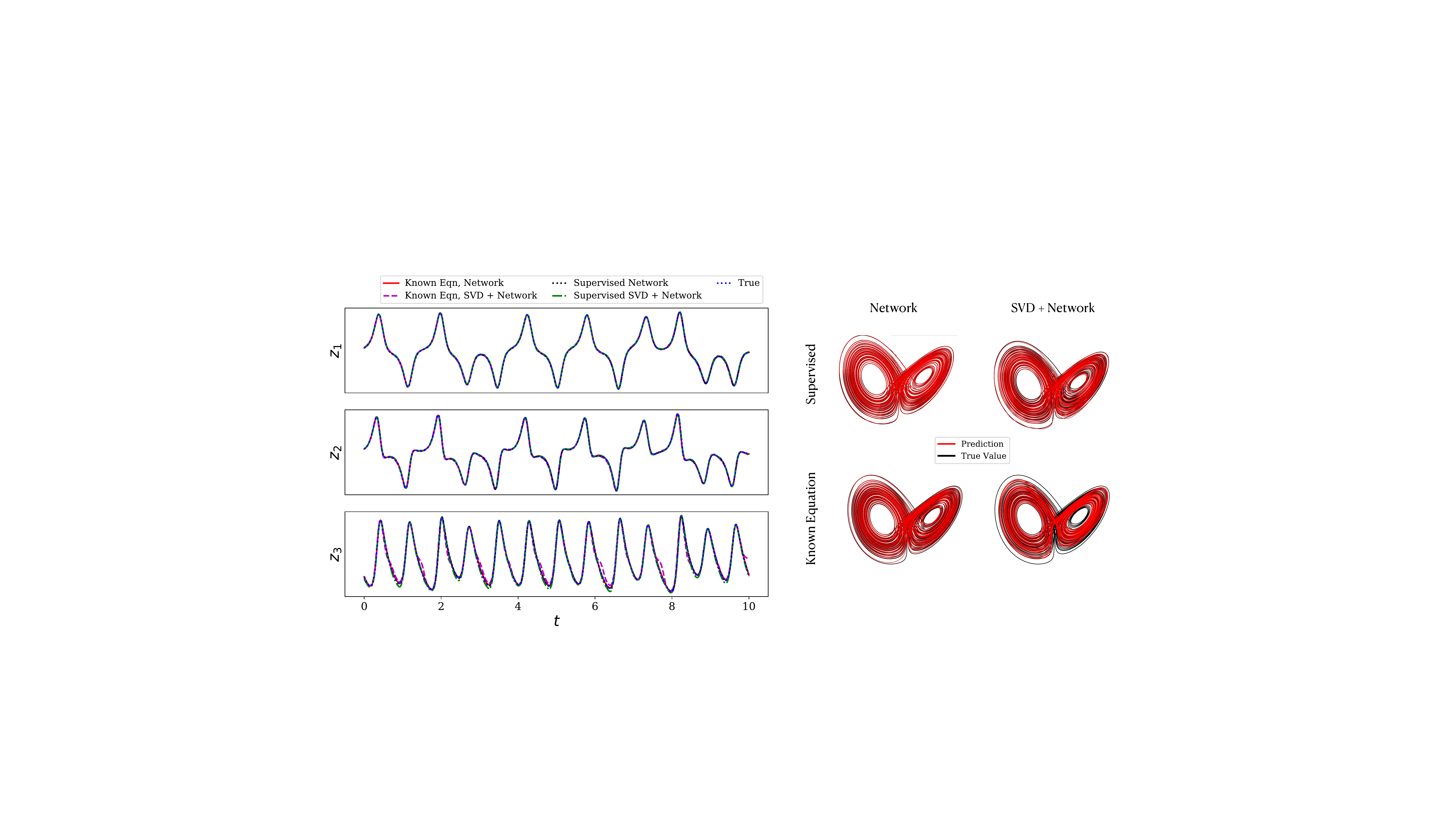}
\caption{Learning an autoencoder map from delay embedding to the full-state attractor, either with a known $\mathbf{x}$ (supervised) or with a known full-state model $\mathbf{f}$ (known equation).  We investigate the performance with and without using an intermediate SVD layer. The parameters are $n=128$, $p=10$, and $m=3$, using $3$ hidden layers in both the encoder and decoder. The computational advantage of using SVD comes at the cost of lower prediction accuracy.}
\label{fig:svdx2z}
\end{center}
\end{figure}

Figure~\ref{fig:svdx2z} shows that sub-problems \ref{subprob:1} and \ref{subprob:2} can be solved with high accuracy, when either the full state $\mathbf{x}$ or the dynamics $\mathbf{f}$ are known, respectively.  For small $p=10$, sub-problem \ref{subprob:2} has a large error for large amplitudes of $x_3$ where the trajectory switches lobes. 
This example also explores performance with and without the use of an intermediate SVD step.  

\begin{figure}[t]
\begin{center}
\includegraphics[width=16cm]{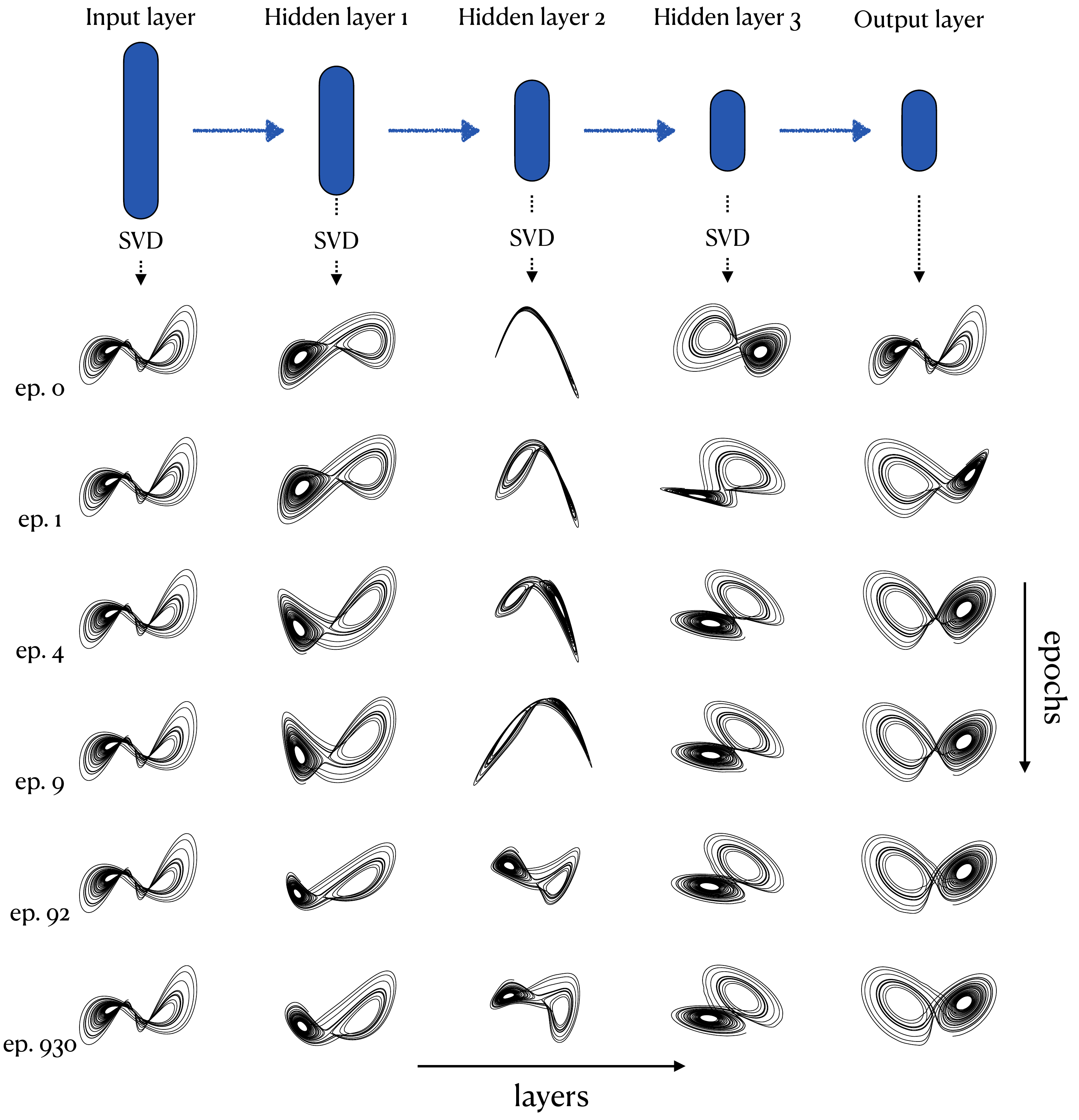}
\caption{Evolution of the attractor as a function of epochs and deep network layers for the known equation case through the encoder. The three linear dominant modes of every network layer are shown for visualization. Columns correspond to layers row to selected epochs. At epoch 0, the network is initialized to map the embedded data $\mathbf h$ to the three dominant modes of $\mathbf H$ at the output layer.}
\label{fig:h2z_viz}
\end{center}
\end{figure}

The diffeomorphic transformations ${\bm \phi}$ and ${\bm \psi}$ are approximated in consecutive nonlinear stages through the hidden layers of the autoencoder.
Figure~\ref{fig:h2z_viz} illustrates the evolution of the attractor as a function of network layers (column-wise) and training epochs (row-wise) when the analytical model is known and fixed in the SINDy coefficients $\mathbf \Xi$.
We use the three dominant modes of every layer for visualization and initialize the network to map $\mathbf{h}$ to $\mathbf v=[v_1, v_2, v_3]$, where $\mathbf v$ are the three dominant right singular vector modes of the Hankel matrix $\mathbf H$.
This initialization of ${\bm \phi}$ and ${\bm \psi}$ results in a solution that is closer to the optimal solution than random initialization, particularly by making them smooth functions. 
This is a form of transfer learning in the context of data-driven modeling.

\begin{figure}[t]
\begin{center}
\includegraphics[width=16cm]{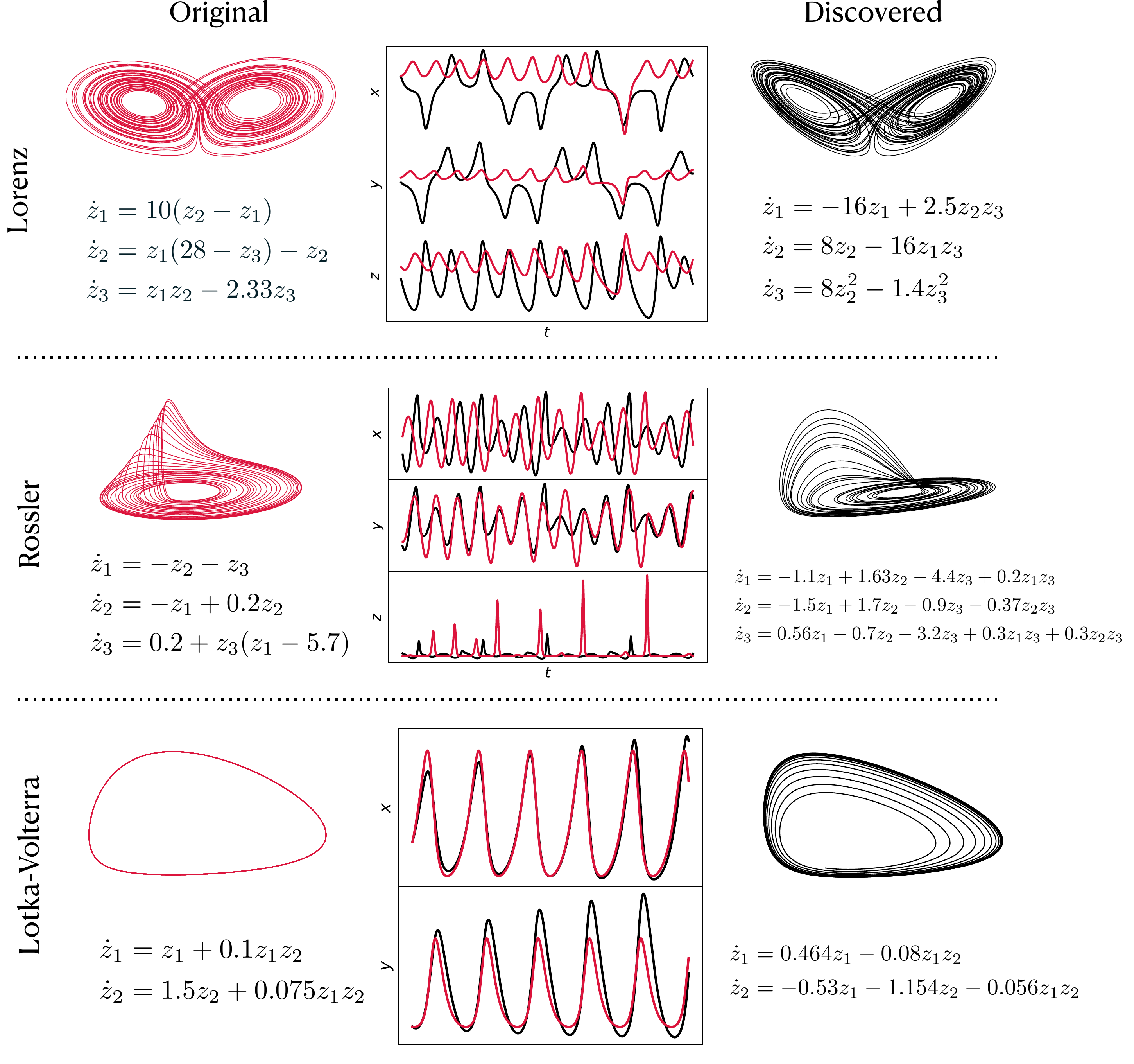}
\caption{Results for Lorenz, R\"ossler, and Lotka Volterra systems.}
\label{fig:summary-fig-1}
\end{center}
\end{figure}

Initializations of $\mathbf \Xi$ with large coefficient perturbations (sampled from a normal distribution $\mathcal N(\xi_{ij}, 20)$) around the original coefficients of $\mathbf f$ consistently lead to Lorenz-like chaotic attractors with a two-lobe structure. 
However, the convergence of the algorithm for random initializations of $\mathbf \Xi$ is sensitive to hyperparameters and does not always give rise to a model that has a similar butterfly structure as the original Lorenz attractor.
This sporadic convergence indicates that the optimizer is prone to over-fitting and getting stuck in local minima, an issue that is commonly observed in deep learning.
We find that applying a separate SINDy optimization on $\mathbf z$ every $k$ epochs (with $k\approx20-40$) improves sparsity and efficiency significantly, while nudging the fitting parameters just enough to avoid local minima.
For a given optimal set of weighting coefficients $\bm \lambda$, we discover a sparse system of equations 
\begin{subequations}\label{eq:disc-eqn-0}
\begin{align}
    \dot z_1 &= -16z_1 + 2.5z_2z_3 \\
    \dot z_2 &= 8 z_2 - 16 z_1z_3\\
    \dot z_3 &= 8 z_2^2 - 1.4 z_3^2,
\end{align}
\end{subequations}
whose corresponding attractor is similar to the original system, as shown in Fig.~\ref{fig:summary-fig-1}. 
The discovered latent representation of the time-delay embedded data is a almost a scaling and translation transform of the original attractor.
This discrepancy is expected from a method that considers linear transforms and rotation of a given dynamical systems to be equally valid diffeomorphic solutions.
Note that this Lorenz-like system is more sparse than the original system of equations, containing 6 instead of 7 terms, but with 4 nonlinear terms instead of 2.
The coefficients of equation \ref{eq:disc-eqn-0} also appear to be involve 3 system parameters ($a = 8$, $b=2.5$ and $c=1.4$), similar to the original system with parameters $\rho$, $\beta$ and $\sigma$. 

\begin{figure}[t]
\begin{center}
\includegraphics[width=16cm]{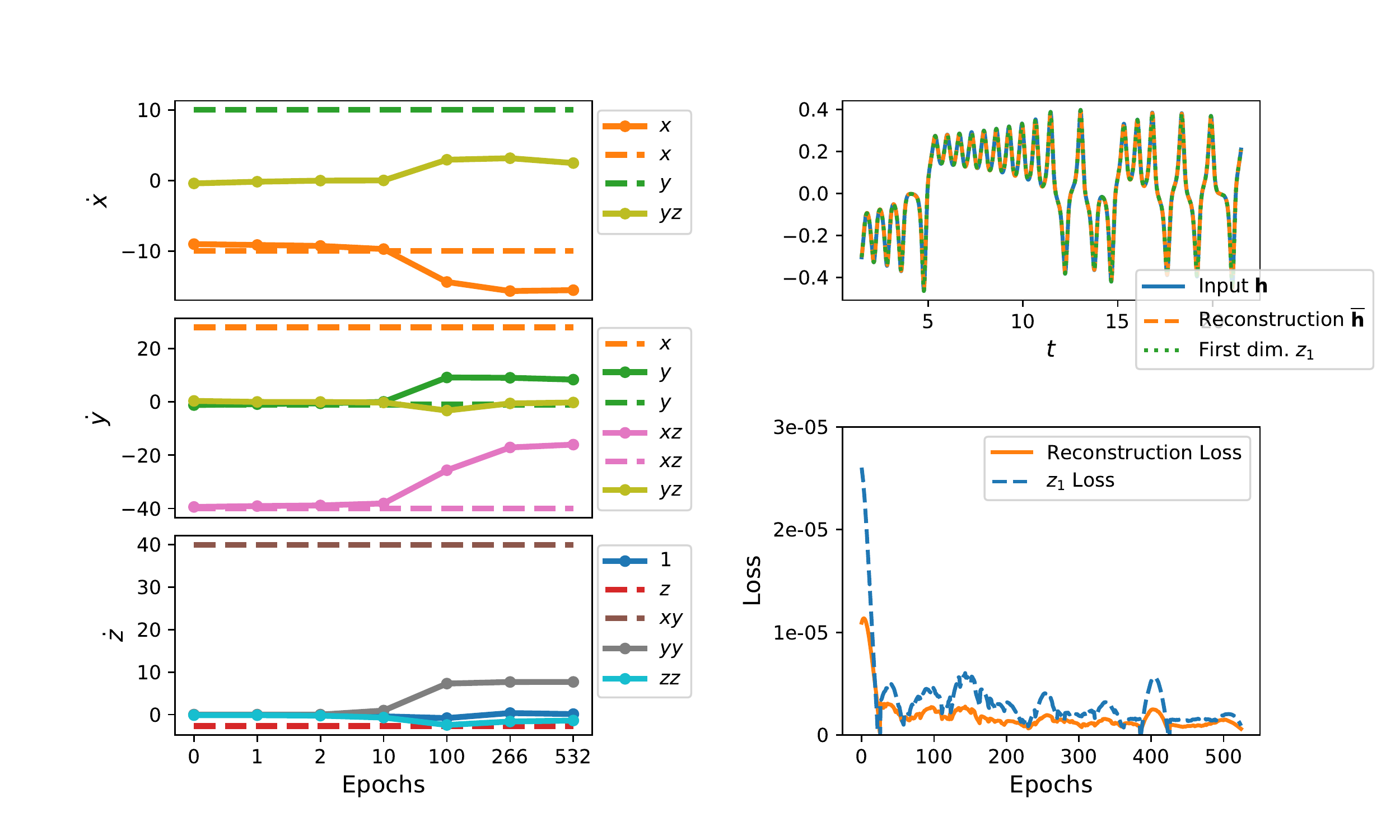}
\caption{Coefficients evolution as a function of epochs, where the dashed lines are the coefficients of the original system and the solid lines are the evolution of the discovered coefficients (left), $z_1$ loss and reconstruction loss as a function of epochs (bottom right), and input reconstructions as a function of time (top right).}
\label{fig:lorenz_epoch_evol}
\end{center}
\end{figure}

Figure ~\ref{fig:lorenz_epoch_evol} shows the evolution of the coefficients as a function of time, the autoencoder reconstruction comparing $\mathbf{h}$ and $\bar{\mathbf{h}}$ (optimized through equation ~\eqref{eq:auto-loss}), and a comparison between the first components of $\mathbf z$ and $\mathbf{\hat{h}}$ (optimized through equation ~\eqref{eq:z1loss}). The input $\mathbf h$ is nearly perfectly reconstructed at the output of the autoencoder and in the first dimension of the latent variable, with loss below $10^{-5}$. The coefficients of the discovered model diverge from the original coefficients, although with a similar two-lobe attractor dynamics.

\subsection{Other systems}
We now test the algorithm on the Lotka-Volterra model and the R\"ossler system, as well as on a challenging example of a video of a chaotic waterwheel experiment.

\paragraph{Lotka Volterra.} The Lotka-Volterra equations describe the interaction of multiple species with a predator-prey relationship. In the simplest case, a system of equations that quantifies the number of a predator $x$ and prey $y$ is given by
\begin{subequations}
\begin{align}
    \dot x &= ax + bxy \\
    \dot y &= cy + dxy,
\end{align}
\end{subequations}
where $a$, $b$, $c$, and $d$ are parameters that depend on the death rate and reproduction rate of both species. In this case, the attractor is a family of limit cycles that depend on the input parameters and the initial condition. 
The discovered models are structurally similar as shown in Fig.~\ref{fig:summary-fig-1}.

\paragraph{R\"ossler system.} The R\"ossler system is a three-dimensional system of differential equations, commonly used for studying chaotic systems, given by
\begin{subequations}
\begin{align}
    \dot x_1 &=  -x_2 - x_3\\
    \dot x_2 &=  -x_1 + ax_2 \\
    \dot x_3 &= a + x_3(x_1 - b),
\end{align}
\end{subequations}
where $a$, $b$ are input parameters. The attractor is known for its intermittently bursting third dimension. Figure~\ref{fig:summary-fig-1} shows that the discovered model is less sparse than the original system; however, the bursting effect in the third dimension is accurately captured.

\begin{figure}[t]
\begin{center}
\includegraphics[width=17cm]{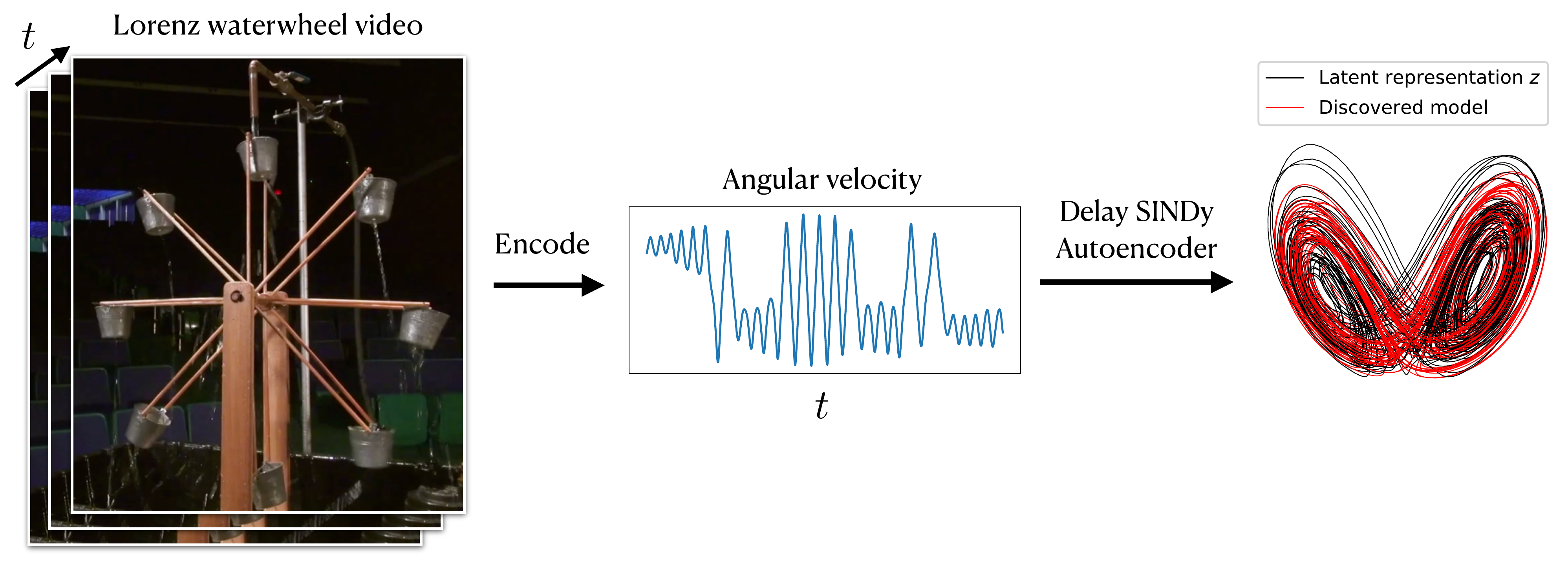}
\caption{Lorenz waterwheel discovery from video~\cite{lorenzww_video}. An OpenCV line detection algorithm is used to find the angle of the spokes and a k-means algorithm is used for clustering lines that detect the same spoke. The angles of the lines are post-processed to deduce one angular velocity $\omega$ for each frame, which is used as an input measurement variable.}
\label{fig:lorenzww}
\end{center}
\end{figure}

\begin{figure}[t]
\begin{center}
\includegraphics[width=13cm]{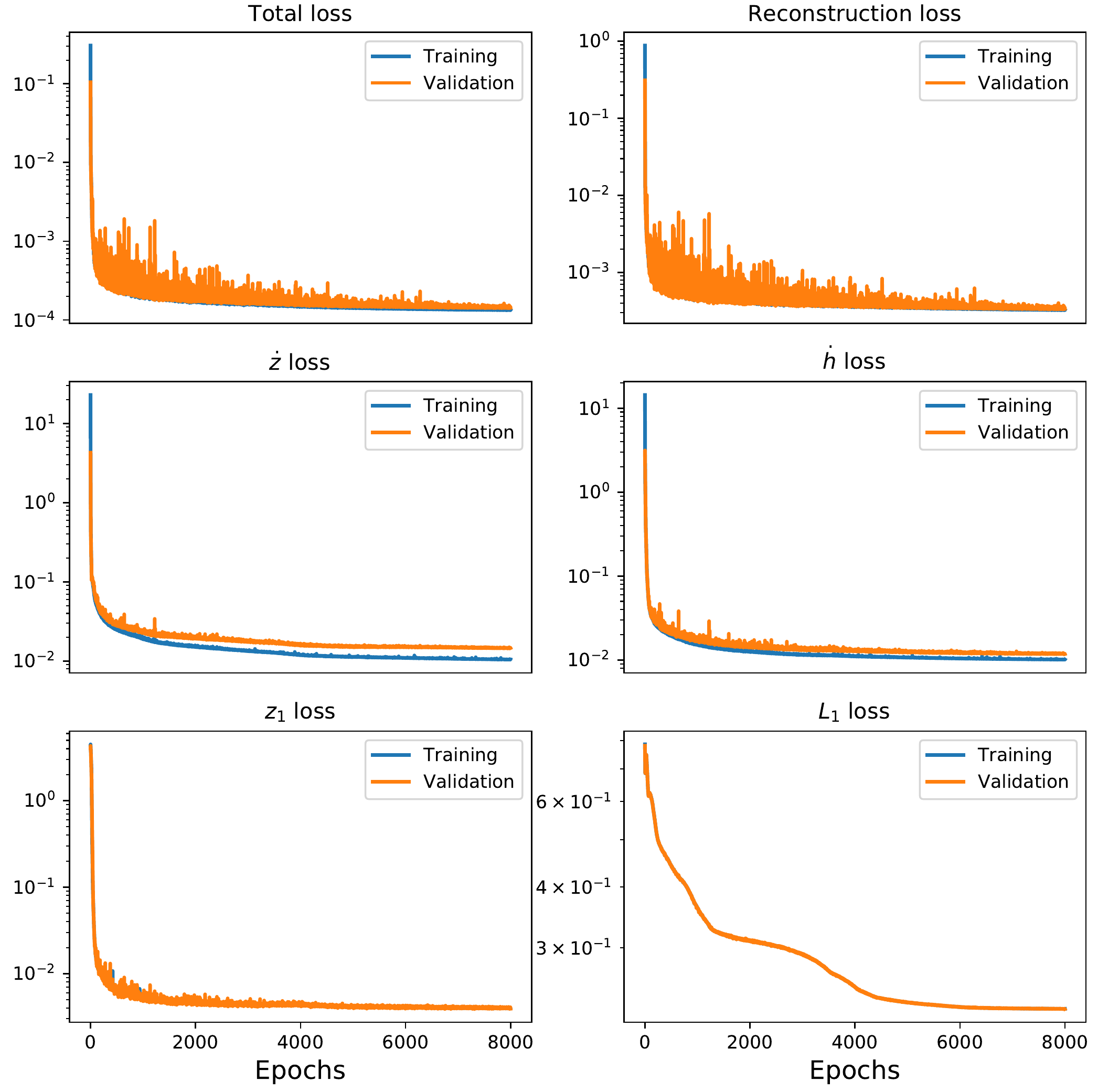}
\caption{Evolution of the losses as a function of epochs for the Lorenz waterwheel case.}
\label{fig:loss_evolution}
\end{center}
\end{figure}

\paragraph{Chaotic waterwheel.} Finally, we use video footage of the Lorenz waterwheel experiment to discover a latent representation and its corresponding system of equations, as shown in Fig.~\ref{fig:lorenzww}. The video is encoded into a time series using the Hough transform to detect the spokes of the wheel, a k-means clustering algorithm to group lines that identify the same spoke, which are post-processed to calculate the angular velocity of the wheel.
A latent representation of the Lorenz-like dynamics of the angular velocity is discovered by the delay SINDy autoencoder, and third degree polynomial features are required in the model for chaotic dynamics (see Appendix \ref{app:lorenzww}). The rough attractor shape and chaotic dynamics are both reproduced. The evolution of the losses as a function of epochs is shown in Fig.~\ref{fig:loss_evolution}.

\section{Discussion}\label{sec:discussion}
This work has shown that it is possible to discover a sparse, interpretable, nonlinear dynamical system from incomplete or partial measurements using a combination of data augmentation through time-delay embedding, nonlinear dimensionality reduction and coordinate transformation with deep neural network autoencoders, and model identification with SINDy. 
Thus, this work begins to resolve a forty year old open challenge, which began with Takens' embedding theorem, to explicitly learn a diffeomorphism back from delay coordinates to a coordinate system where the dynamics may be efficiently represented.   
We also highlight the importance of choosing the proper hypothesis class, as constrained by the relevant loss function. 
While the resulting analytical reconstructions do not always match the original system from which the data was generated, the discovered latent variables and their underlying models are sparse, sharing key structural features with the original system when it is known. 
Surprisingly, we discover Lorenz-like models that are more sparse than the original Lorenz equations from 1D measurements.  
Our models have only six terms, while the original Lorenz equations have seven terms; however, additional sparsity comes at the cost of a higher number of quadratic terms.
The method's ability to discover models that are qualitatively and quantitatively similar to the original system, when the differential equation is initialized with perturbations around the original coefficients, suggests that the optimizer is prone to over-fitting and becoming stuck in local minima, resulting in large estimation errors. 
Designing and adding appropriate regularization terms to the loss function to solve this issue remains an open question and is the subject of further investigation.

It is unknown whether or not there exists a loss function for which the original system of equations can be exactly recovered from one-dimensional measurements and random fitting parameter initialization. 
There is always the possibility of adding more physics-informed constraints to the loss function, such as known problem-specific symmetries, to help the optimizer find the desired model. 
Furthermore, the diffeomorphic transform between the delay embedding and the full-state attractor can be further enforced by using a bijective neural network, i.e. an invertible neural network, that honors the one-to-one map by construction. 
In addition, the SINDy algorithm used in this study is relatively basic in its use of the $L_1$ loss for enforcing sparsity, and many improvements have been suggested in the past few years~\cite{deSilva2020JOSS,kaptanoglu2021pysindy}, which can be integrated into the proposed delay SINDy autoencoder framework. Particularly, using a probabilistic~\cite{fasel2021ensemble} or a weak~\cite{Schaeffer2017prsa,gurevich2019robust,Reinbold2020pre,reinbold2021robust} SINDy algorithm would certainly improve on the results obtained from real-world noisy data such as the Lorenz waterwheel video.

Crucially, this study proposes a general framework for discovering multi-dimensional models from partial, low-dimensional data, motivating many theoretical and numerical investigations into the combined use of time-delay embedding, SINDy, and autoencoders for data-driven modeling. Future work will involve exploring various deep network architectures, such as convolutional autoencoders, for discovering physics from high-dimensional video data.

\section*{Acknowledgments}
The authors acknowledge support from the Army Research Office (ARO W911NF-19-1-0045) and the National Science Foundation AI Institute in Dynamic Systems (Grant No. 2112085).

\appendix

\section{Delay embedding hyperparameters}\label{app:embedding-hyper}
\begin{figure}[t]
\begin{center}
\vspace{0.2in}
\begin{overpic}[trim=0 0 0 40, clip,width=14cm]{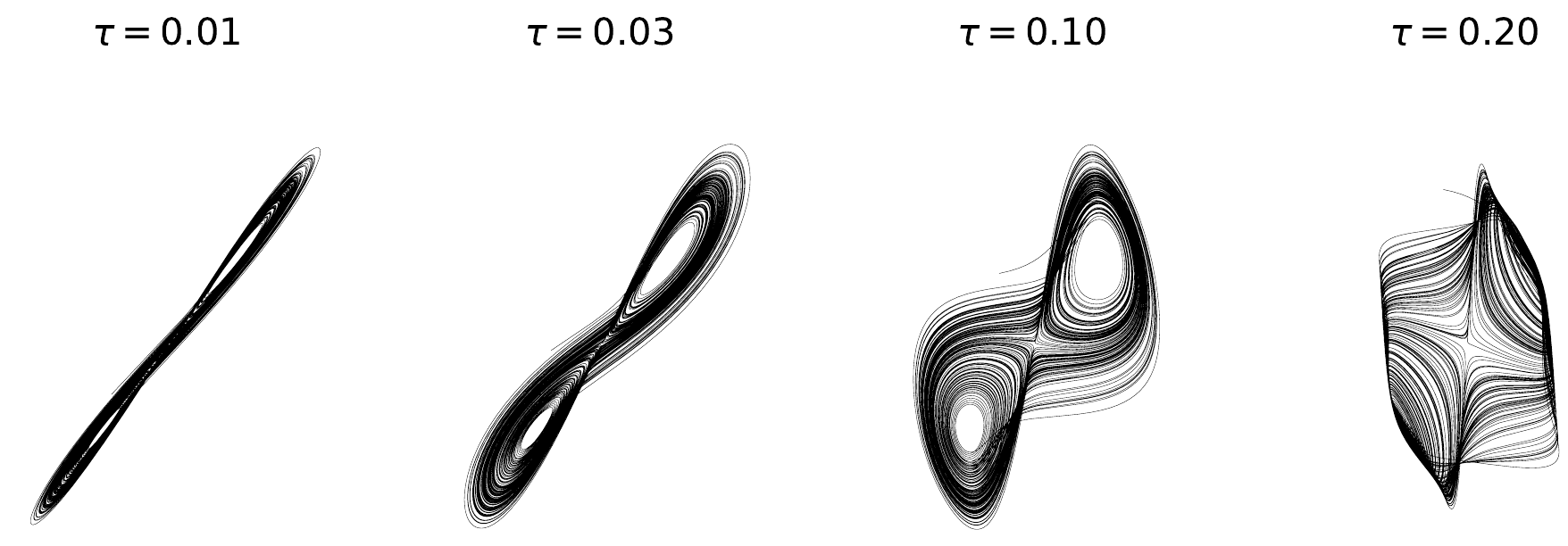}
\put(6,28){$n\tau=0.01$}
\put(32.5,28){$n\tau=0.03$}
\put(61,28){$n\tau=0.10$}
\put(87,28){$n\tau=0.20$}
\end{overpic}
\caption{The delay time increment $n\tau$ is chosen to optimally unfold the attractor in the embedding space. In the case of the Lorenz attractor, $n\tau \approx 0.1$ seems to be the most appropriate choice.}
\label{fig:time-delay-comp}
\end{center}
\end{figure}
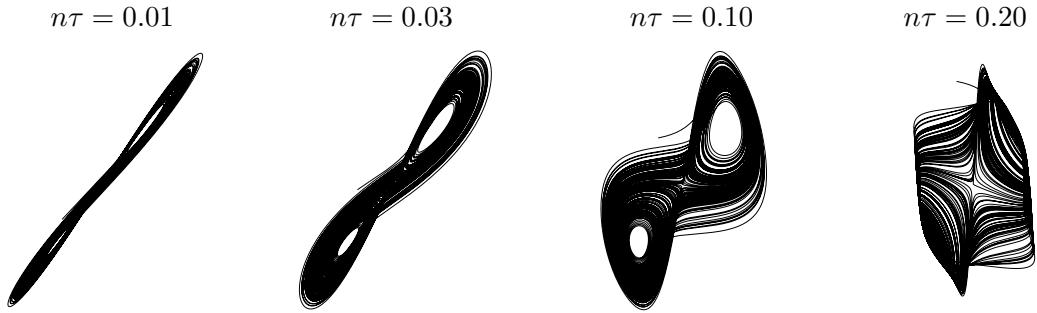

The time delay increment $n\tau$, where $\tau$ is the sample time and $n$ is the number of delays included in the Hankel matrix, is typically chosen to make the attractor occupy (or ``unfold'' in) as much of the embedded phase space as possible. 
Methods for choosing $n\tau$ using measures of autocorrelation or mutual information~\cite{fraser1989information} to maximize the phase-space unfolding have been studied extensively~\cite{kennel1992method, ma2006selection, wallot2018calculation}. 
Sampling strategies have also been studied for discovering physics from data~\cite{champion2019data}.
In this study, we choose the embedding dimension $n$ and the time delay increment $n\tau$ to match that chosen in HAVOK~\cite{brunton2017chaos}. 
A comparison of the unfolding of the attractor based on $n\tau$ alone in figure~\ref{fig:time-delay-comp} shows that the optimal unfolding is for $n\tau \approx 0.1$.

\begin{figure}[t]
\begin{center}
\includegraphics[width=17.5cm]{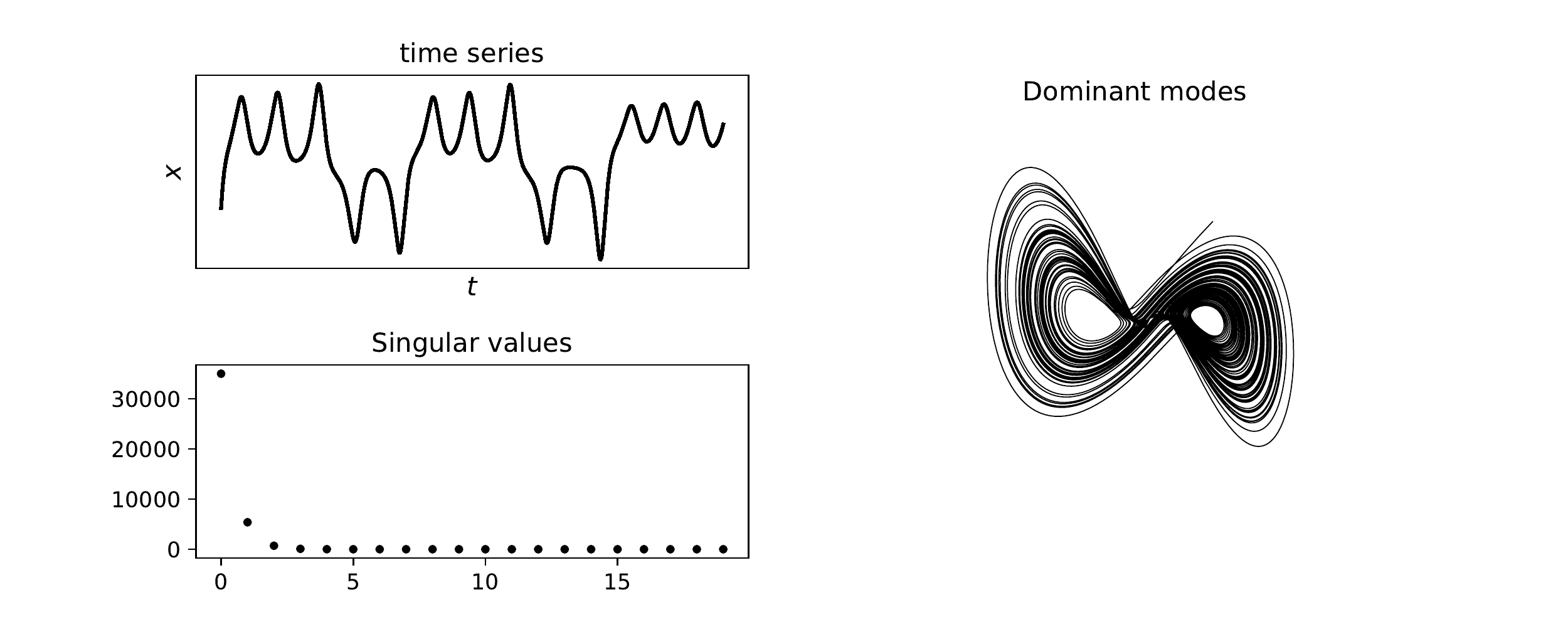}
\caption{The first 3 SVD modes of the Hankel matrix reflect the two-lobe structure and capture a majority of the variance.}
\label{fig:hankel}
\end{center}
\end{figure}

When finding the dominant modes of the Hankel matrix, similar considerations for the dimension $n$ have to be taken into account. Figure~\ref{fig:hankel} shows that the first three dominant modes are representative of the majority of the variance in the data. This helps guide the choice of the latent dimension $m$ when it is not known a priori. More generally, standard techniques for determining the number of autoencoder latent variables may be used before adding the SINDy losses.  

\begin{figure}[t]
\begin{center}
\includegraphics[width=17cm]{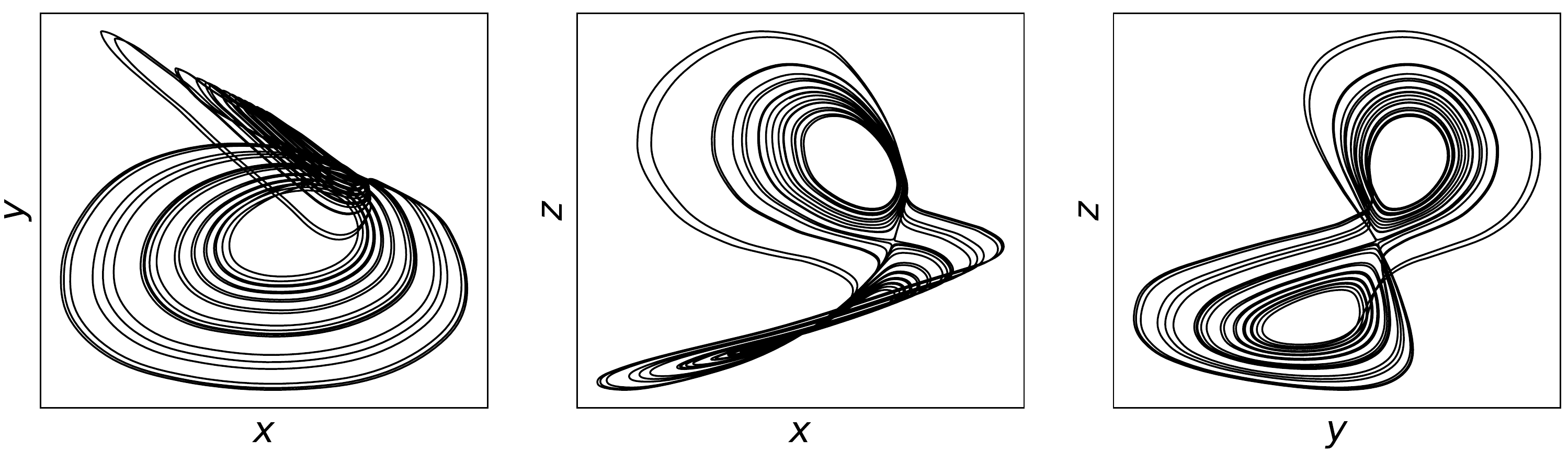}
\caption{Autoencoder latent representation with only a reconstruction loss (without SINDy). Error is $\varepsilon \sim 7\cdot 10^{-7}$.  This coordinate system is unlikely to yield a sparse model.}
\label{fig:x2z2x}
\end{center}
\end{figure}

It is reasonable to ask if the the learning of coordinates and identification of models can be separated into sequential learning steps.  Fig.~\ref{fig:x2z2x} shows the latent representation $\mathbf z$ from an autoencoder without SINDy constraints.  
Because the embedding is non-unique, the autoencoder will typically fail to capture essential symmetries from the original system, making it difficult or impossible to achieve sparse models, thus motivating the combined SINDy-autoencoder approach. 

\section{Lorenz waterwheel model}
\label{app:lorenzww}
The discovered Lorenz waterwheel attractor shown in Fig.~\ref{fig:lorenzww} has the same two-lobe structure as the original Lorenz system. However, the sparsity of the data, the measurement noise that arises from the line detection algorithm, the small frame rate of the video (24 frames/second) and its limited time span make the sparse identification of the differential equation a challenging task. After optimizing over hyperparameters, a third order polynomial dictionary (with $n=32$ and a $4$ layer encoder/decoder) is found to recreate the chaotic dynamics of the Lorenz system. The discovered coefficients are shown in table~\ref{tab:coefficients}, where the differential equation is written as
\begin{equation}
    \dot z_i(t) = \sum_j \xi_{ij} \theta_j.
\end{equation}

A sparser and more generalizing analytical model can most likely be found by collecting more data and improving the algorithm that maps the video to the angular velocity. In future studies, we will explore machine learning techniques (e.g. using a CNN) to incorporate the video to angular velocity mapping as part of the SINDy autoencoder algorithm.

{\renewcommand{\arraystretch}{1.3}
\begin{table}[h]
\hspace{-.2in}
\scriptsize
 \begin{tabular}{|c||c|c|c|c|c|c|c|c|c|c|c|c|c|c|c|c|c|c|c|c|} \hline
        $\theta$& $1$ & $z_1$ & $z_2$ & $z_3$ & $z_1^2$ & $z_1z_2$ &
        $z_1z_3$ & $z_2^2$ & $z_1z_2$ & $z_3^2$ & $z_1^3$ & 
        $z_1^2 z_2$ & $z_1^2 z_3$ & $z_1z_2^2$ & $z_1z_2z_3$ & 
        $z_1z_3^2$ & $z_2^3$ & $z_2^2z_3$ & $z_2z_3^2$ & $z_3^3$ \\
        \hline \hline
         $\dot z_1 $ & -0.12  &  1.70 &   0.04 & &  
           & -1.79 &      &  0.16 &       &  
            &       &      &    &     &  &
            & -0.02 &  0.03 & -0.37 &         \\
         $\dot z_2 $ & -1.05  &  0.05 & -0.28 &   & 
             0.12 &    & 1.50 &  0.60 &    &    &    &     &    &    &-0.10&   &     &  0.18 & -0.33 &\\
        $\dot z_3 $ & 0.08 & -0.12 &  0.30 &  0.07 &  0.08 &
            -0.29 & -0.22 &  0.35 & -0.55 & &
             &  0.06 &  & -0.07 &  0.20
              &  0.03 & -0.25 &  0.28 & & \\ \hline
    \end{tabular}
    \caption{Discovered coefficients of the Lorenz waterwheel system.}
    \label{tab:coefficients}
\end{table}
}
\bibliographystyle{unsrt}
\begin{spacing}{.88}
\small{
\setlength{\bibsep}{4.pt}
\bibliography{references.bib}
}
\end{spacing}

\end{document}